\documentclass[journal,twoside,web]{ieeecolor}
\usepackage{generic}
\usepackage{cite}
\usepackage{amsmath,amssymb,amsfonts}
\usepackage{algorithmic}
\usepackage{algorithm}
\usepackage{graphicx}
\usepackage{textcomp}
\usepackage[caption=false,font=normalsize,labelfont=sf,textfont=sf]{subfig} 
\usepackage{array}
\usepackage{booktabs}  
\usepackage{multirow}  
\usepackage{stfloats}
\usepackage{url}

\def\BibTeX{{\rm B\kern-.05em{\sc i\kern-.025em b}\kern-.08em
    T\kern-.1667em\lower.7ex\hbox{E}\kern-.125emX}}
\begin{document}
\title{Task-guided Spatiotemporal Network with Diffusion Augmentation for EEG-based Dementia Diagnosis and MMSE Prediction}

\author{Xiaoyu Zheng\textsuperscript{a}, Xu Tian\textsuperscript{a}, Bin Jiao\textsuperscript{a}, Kunbo Cui, Hanhe Lin, Lu Shen\textsuperscript{*}, and Jin Liu\textsuperscript{*}
\thanks{This work was supported in part by the National Natural Science Foundation of China (No. 62472450), the Science and Technology Innovation Program of Hunan Province (No. 2025RC3020), the Central South University Innovation-Driven Research Programme (No. 2023CXQD018), the High Performance Computing Center of Central South University, and the Fundamental Research Funds for the Central Universities of Central South University (No. 2026ZZTS0038).}
\thanks{Xu Tian and Jin Liu are with the Hunan Provincial Key Lab on Bioinformatics, School of Computer Science and Engineering, Central South University, Changsha 410083, China.}
\thanks{Xiaoyu Zheng, Kunbo Cui, and Jin Liu are with the Xinjiang Engineering Research Center of Big Data and Intelligent Software, School of Software, Xinjiang University, Urumqi 830091, China.}
\thanks{Hanhe Lin is with School of Science and Engineering, University of Dundee, DD1 4HN Dundee, United Kingdom.}
\thanks{Bin Jiao and Lu Shen are with the Department of Neurology, Xiangya Hospital, Central South University, Changsha 410083, China.}
\thanks{\textsuperscript{a} Xiaoyu Zheng, Xu Tian, and Bin Jiao contribute equally to this paper.}
\thanks{\textsuperscript{*} Corresponding authors: Jin Liu and Lu Shen (e-mail: liujin06@csu.edu.cn; shenlu@csu.edu.cn).}}


\maketitle

\begin{abstract}
Patients with dementia typically exhibit cognitive impairment, which is routinely assessed using the Mini-Mental State Examination (MMSE). Concurrently, their underlying neurophysiological abnormalities are reflected in Electroencephalography (EEG), providing a basis for joint modeling. 
However, traditional multi-task approaches suffer from feature entanglement, which leads to inter-task interference when handling heterogeneous objectives.
To address this challenge, we propose a task-guided spatiotemporal network (TGSN) with diffusion augmentation for EEG-based dementia diagnosis and MMSE prediction. 
Specifically, TGSN integrates a multi-band feature fusion module to capture complementary spectral information from EEG. Meanwhile, a pre-trained data augmentation module utilizing a diffusion process is introduced toincrease sample diversity. To model the complex spatiotemporal patterns of EEG, we propose a gated spatiotemporal attention module that captures long-range spatial dependencies and temporal dynamics.
Moreover, we design a task-guided query module to achieve task-specific feature extraction, thereby mitigating task interference.
The effectiveness of TGSN is evaluated on the XY02 dataset. 
Experimental results demonstrate that the proposed network outperforms several state-of-the-art methods, achieving classification accuracies of 97.78\% for Alzheimer's Disease (AD)/Frontotemporal Dementia (FTD) and 83.93\% for AD/FTD/Vascular Cognitive Impairment (VCI), which exceed the best baselines by 16.39\% and 8.28\%, respectively. In parallel, it reduces the RMSE for MMSE prediction to 1.93 and 2.38, achieving significant error reductions of 1.44 and 1.43 compared to the best baselines.
Additionally, validation on the DS004504 dataset demonstrates strong cross-dataset generalization. 
These results indicate that TGSN provides an effective and robust solution for multi-task EEG-based dementia modeling. The source code will be made publicly available upon acceptance.
\end{abstract}


\section{Introduction}
\IEEEPARstart{D}{ementia} is a syndrome encompassing multiple neurodegenerative disorders that lead to severe cognitive impairment \cite{arvanitakis2019diagnosis}. Among these, Alzheimer’s Disease (AD), Frontotemporal Dementia (FTD), and Vascular Cognitive Impairment (VCI) are the most common types \cite{ng2021distinct, maheshwari2024navigating}. Due to overlapping clinical manifestations, differentiating among these conditions remains challenging yet important \cite{goodwin2025comparing}. In clinical practice, the Mini-Mental State Examination (MMSE) is widely used for screening and evaluating cognitive impairment caused by dementia due to its simplicity and efficiency \cite{arevalo2021mini}. The MMSE reflects the overall cognitive function of patients and provides important support for dementia diagnosis \cite{creavin2016mini}. Therefore, jointly modeling dementia diagnosis and MMSE prediction can leverage their intrinsic correlation, thereby improving diagnostic accuracy and reliability.

Electroencephalography (EEG) is a non-invasive technique that records the electrical activity of the cerebral cortex with high temporal resolution and has been extensively investigated for the analysis of dementia and cognitive impairment \cite{gami2025emerging}. By providing millisecond-level temporal dynamics and spatial information of brain activity, EEG enables effective characterization of neurological abnormalities associated with cognitive impairment \cite{song2021transformer}. 
EEG is commonly decomposed into multiple frequency bands, including Delta, Theta, Alpha, Beta, and Gamma, each reflecting distinct functional states of the brain and providing valuable biomarkers for dementia diagnosis \cite{klassen2011quantitative,wei2025multi}.

\begin{figure}[!htbp]
\centering
\includegraphics[width=0.45\textwidth]{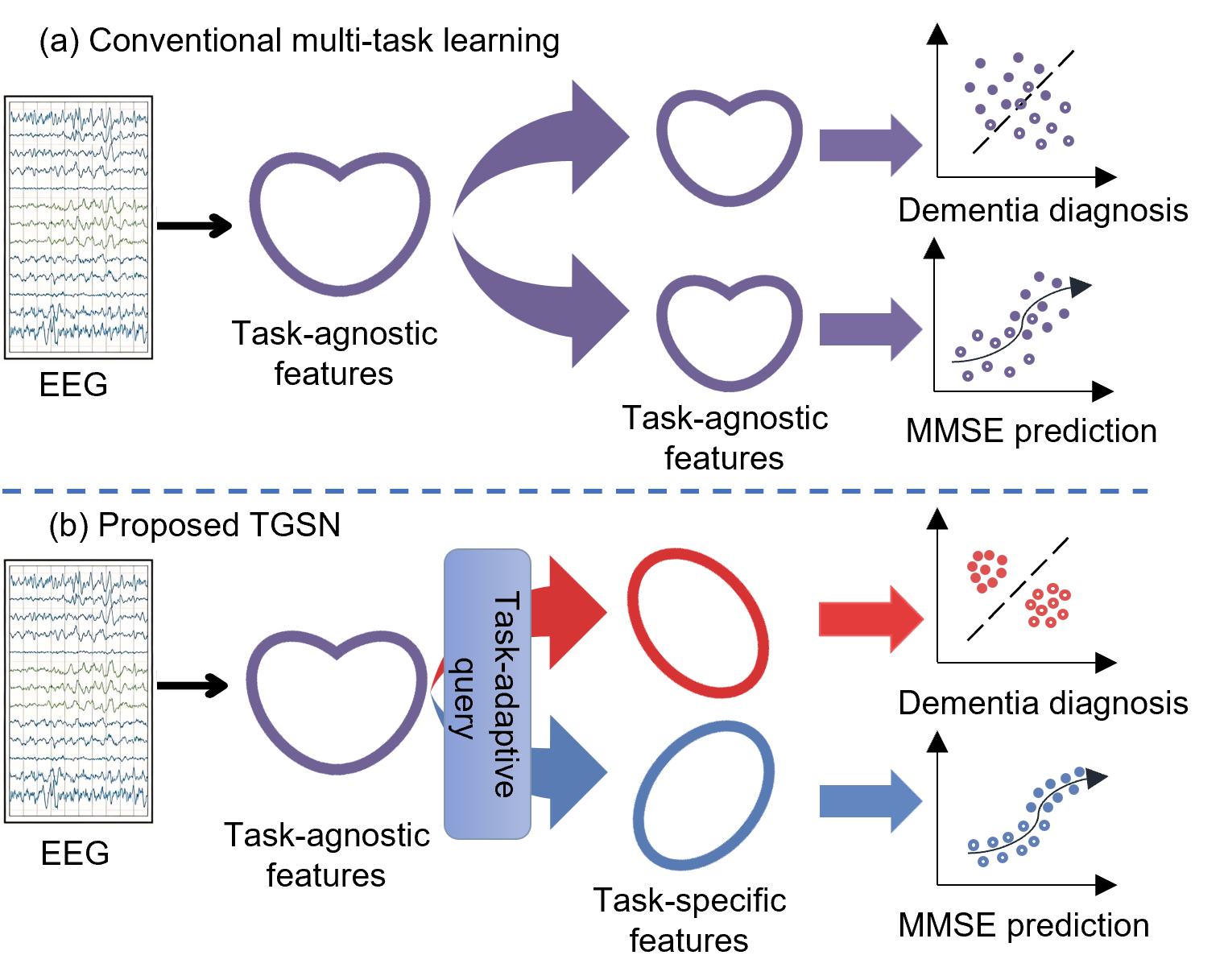}
\caption{\textbf{Illustration of the challenge and our solution.} (a) Conventional multi-task learning suffers from feature entanglement when handling heterogeneous tasks. (b) The proposed TGSN employs a task-guided query module for task-specific feature extraction, thereby mitigating inter-task interference.} 
\label{challenge}
\end{figure}
Although machine learning has advanced EEG-based dementia diagnosis \cite{miltiadous2021alzheimer,safi2021early,oltu2021novel,zheng2024multi} and MMSE prediction \cite{luo2022prediction,qiao2022ranking}, most studies treat these intrinsically correlated tasks independently. While multi-task learning (MTL) can leverage shared neurophysiological patterns \cite{li2022emotion,li2023mtlfusenet}, the joint optimization of discrete diagnostic labels and continuous cognitive scores often causes feature entanglement, leading to inter-task interference.
To address this challenge, we propose a task-guided spatiotemporal network (TGSN) with diffusion augmentation for EEG-based dementia diagnosis and MMSE prediction, as shown in Fig.~\ref{challenge}. The proposed network integrates multi-band feature fusion, pre-trained data augmentation, and gated spatiotemporal attention modeling to extract feature representations from EEG. Furthermore, it introduces a task-guided query module, which effectively suppresses inter-task interference through task-specific feature extraction. The main contributions of this study are summarized as follows:
\begin{enumerate}
    \item We propose a task-guided spatiotemporal network. By mitigating feature entanglement and enhancing task specificity, it improves the dementia diagnosis and MMSE prediction jointly. 
    \item A pre-trained data augmentation module is introduced to increase sample diversity, while multi-band spatiotemporal features and a task-guided query module are employed to learn discriminative EEG representations.
    \item Extensive experiments demonstrate the effectiveness, robustness, and interpretability of the proposed network, achieving state-of-the-art performance in multiple dementia diagnosis tasks and MMSE prediction.
\end{enumerate}

\begin{figure*}[!htbp]
\vspace{-5mm}
\centering
\setlength{\abovecaptionskip}{-3pt}
\includegraphics[width=0.82\textwidth]{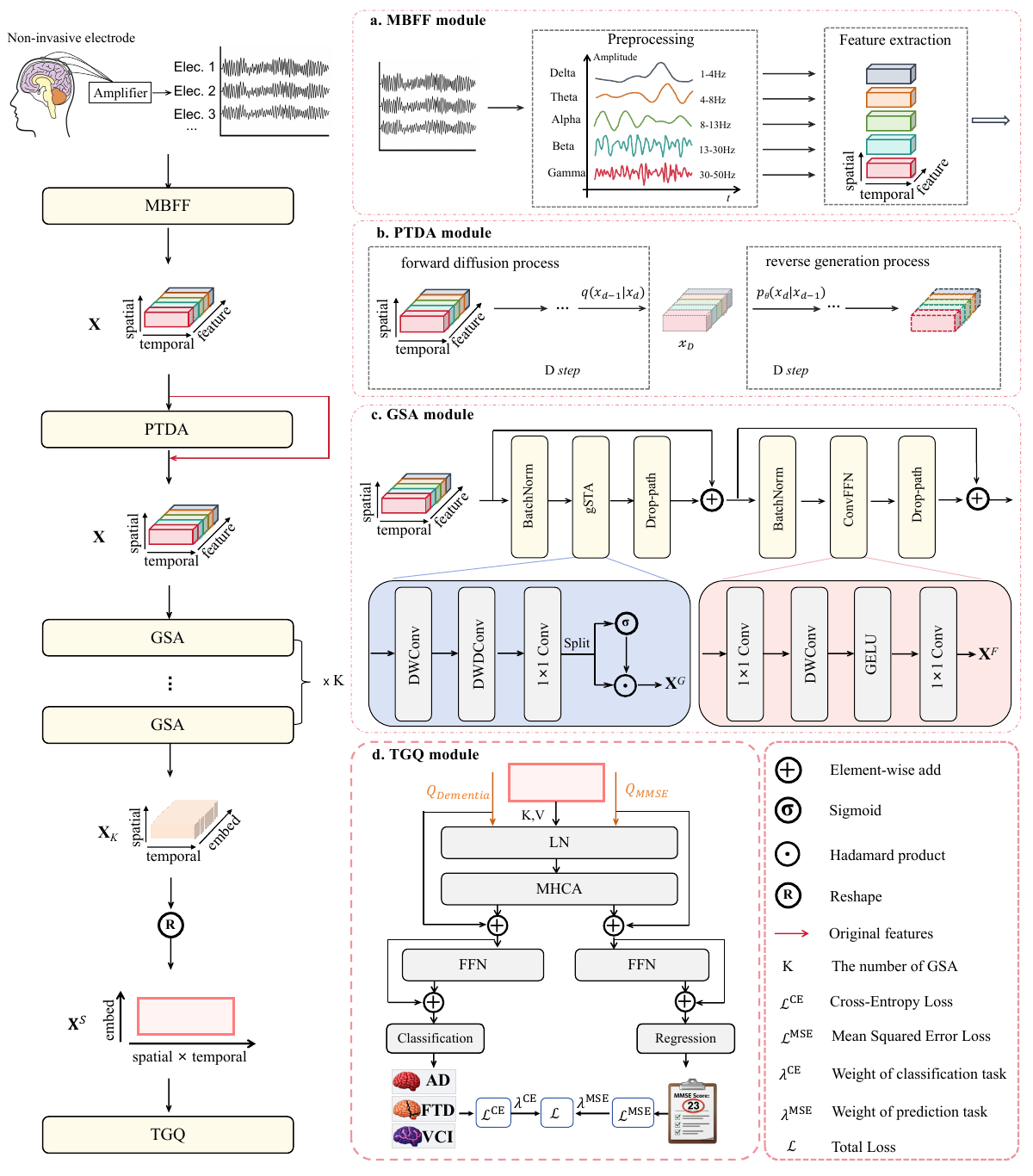}
\caption{\textbf{Framework of the proposed TGSN.} 
\textbf{a. Multi-band feature fusion (MBFF) module}: Performs EEG preprocessing, band decomposition, and feature extraction.
\textbf{b. Pre-trained data augmentation (PTDA) module}: Employs a pre-trained strategy to generate EEG features, enhancing data diversity.
\textbf{c. Gated spatiotemporal attention (GSA) module}: Models long-range spatial dependencies and temporal dynamics from multi-channel EEG representations.
\textbf{d. Task-guided query (TGQ) module}: Mitigates inter-task interference via task-specific feature extraction while jointly optimizing dementia diagnosis and MMSE prediction.} 
\label{model}
\vspace{-10pt}
\end{figure*}

\section{Related work}
\subsection{Single-task learning}
Early EEG-based dementia diagnosis primarily relies on manually extracted features combined with traditional classifiers. Researchers identified biomarkers such as spectral power, entropy, and complexity to distinguish AD from healthy controls or other dementia subtypes \cite{jiao2023neural,zheng2024multi,lee2025diagnosis}. Recognizing that cognitive impairment manifests distinct alterations across frequency bands, subsequent studies introduced multi-band analysis strategies \cite{ieracitano2020novel,araujo2022smart}. Recently, deep learning has significantly advanced EEG-based dementia diagnosis by enabling end-to-end spatiotemporal modeling. Convolutional Neural Networks (CNNs) and Graph Neural Networks (GNNs) have been leveraged to capture complex spatiotemporal dependencies \cite{chen2023multi,kachare2024steadynet,watanabe2024deep,bravo2024spectrocvt,wei2025multi}. Additionally, self-attention mechanisms and the strategic ranking of specific spatiotemporal windows have been widely adopted to selectively focus on disease-relevant brain regions and temporal patterns \cite{sibilano2024understanding,taub2024ranking}.

The MMSE is widely used as a critical quantitative metric for cognitive assessment. While prior research predominantly utilized neuroimaging modalities such as Magnetic Resonance Imaging (MRI) for MMSE prediction \cite{liu2019weakly,li2023machine}, recent studies have shifted towards EEG-based methods due to their high temporal resolution and non-invasiveness. Several methods have been applied to map EEG patterns to cognitive scores \cite{lin2021predicting,sun2024ensemble}. Furthermore, quantitative analysis of EEG rhythmicity has established objective links between neurophysiological signals and cognitive scores \cite{luo2022prediction,qiao2022ranking}.

Although EEG has been extensively applied in studies such as dementia diagnosis and MMSE prediction, the above-mentioned studies have primarily focused on single-task modeling, failing to fully explore shared neurophysiological mechanism underlying these two highly correlated tasks. 

\subsection{Multi-task learning}
MTL is an effective paradigm that exploits shared representations to capture task correlations \cite{vandenhende2021multi}. Given the pathological link between dementia diagnosis and cognitive assessment, recent studies have explored jointly modeling these tasks within a unified framework.
Liu et al. \cite{liu2024multi} demonstrated the benefits of joint learning by modeling feature interactions across dementia diagnosis and MMSE prediction tasks. In the EEG domain, MTL combined with advanced feature extraction strategies has been widely applied to tasks such as emotion recognition and motor imagery \cite{li2022gmss, luo2023dual, wang2025gated}, showing its capability in handling complex spatiotemporal patterns. Inspired by these advances, recent works have introduced MTL into EEG-based dementia analysis, employing Transformer-based architectures and functional brain network modeling to simultaneously optimize dementia diagnosis and cognitive assessment objectives \cite{ilias2022explainable,si2023differentiating}.
Although existing studies have demonstrated the effectiveness of MTL for dementia diagnosis, shared representations across tasks often suffer from feature entanglement and insufficient task specifiability. Therefore, developing a multi-task modeling mechanism that can adaptively balance shared and task-specific representations based on spatiotemporal EEG features remains a key challenge.

\section{Methodology}
In this study, we propose the TGSN, which consists of four core modules as shown in Fig.~\ref{model}: (1) Multi-band Feature Fusion (MBFF) module for multi-band feature extraction; (2) Pre-trained Data Augmentation (PTDA) module to address data scarcity via diffusion-based augmentation; (3) Gated Spatiotemporal Attention (GSA) module for modeling spatiotemporal dynamics; and (4) Task-guided Query (TGQ) module for mitigating inter-task interference via task-specific feature extraction while facilitating the balanced optimization of the two tasks.

\subsection{Multi-band feature fusion module}
\label{sec3.1}

To mitigate noise and extract informative EEG features, EEG is decomposed into frequency bands using band-pass filtering (Delta: 1-4 Hz, Theta: 4-8 Hz, Alpha: 8-13 Hz, Beta: 13-30 Hz, and Gamma: 30-50 Hz) and segmented into non-overlapping $2 s$ epochs. 
We define the EEG as $S \in \mathbb{R}^{B \times C \times T_1 \times T_2}$, where $B, C, T_1$, and $T_2$ denote the number of bands, channels, epochs, and sample points per epoch, respectively. For each frequency band and channel in epoch $t_1$, $S_{b,c,t_1} \in \mathbb{R}^{ T_2}$ is used to extract time- and frequency-domain features to comprehensively characterize neural dynamics \cite{wei2025multi}.

In the time domain, we employ Hjorth parameters and Sample Entropy (SampEn). Hjorth Mobility (HM) and Complexity (HC) quantify the dominant frequency and bandwidth variations of the signal, calculated as:
\begin{equation}
    \mathrm{HM}_{b,c,t_1} = \sqrt{ \frac{\mathrm{Var} (\displaystyle{\mathrm{d}(S_{b,c,t_1})} / {\mathrm{d}t})}{\mathrm{Var} ({S_{b,c,t_1}})}},
\end{equation}
\begin{equation}
    \mathrm{HC}_{b,c,t_1} = \frac{\mathrm{HM}\left({\mathrm{d}(S_{b,c,t_1})}/{\mathrm{d}t}\right)}{\mathrm{HM}\bigl(S_{b,c,t_1} \bigr)},
\end{equation}
where $\mathrm{Var}(\cdot)$ denotes variance operation. SampEn measures the complexity and self-similarity of EEG \cite{xie2011complexity}, defined as follows: 
\begin{equation}
    \mathrm{SampEn}_{b,c,t_1}=-\ln\frac{\mathrm{U}{\bigl(S_{b,c,t_1},m+1,\,r\bigr)}}{\mathrm{V}{\bigl(S_{b,c,t_1},m,\,r\bigr)}},
\end{equation}
where $\mathrm{V}{\bigl(S_{b,c,t_1},m,\,r\bigr)}$ and $\mathrm{U}{\bigl(S_{b,c,t_1},m+1,\,r\bigr)}$ denote the number of template vector pairs of length $m$ and $m+1$ within tolerance $r$, respectively, derived from the EEG $S_{b,c,t_1}$.

In the frequency domain, Welch’s method \cite{welch2003use} is applied to estimate the Power Spectral Density (PSD). To minimize inter-subject variability, we compute the Relative Spectral Density (RSD). We utilize both PSD and RSD as frequency-domain features, the specific formulas are as follows:
\begin{equation}
   \mathrm{PSD}_{b,c,t_1} =   \frac{1}{P} {\sum\limits_{p=1}^{P}} \mathrm{Welch}{(S_{b,c,t_1},p)}, 
\end{equation}
\begin{equation}
    \mathrm{RSD}_{b,c,t_1} = \frac{\mathrm{PSD}_{b,c,t_1}}{\sum\limits_{i=1}^{B} \mathrm{PSD}_{i,c,t_1}},
\end{equation}
where $P$ denotes the number of segments used in Welch’s method, $\mathrm{Welch}{(\cdot,p)}$ denotes the Welch periodogram of the $p$-th segment, and $B$ denotes the total number of frequency bands.
After feature extraction and band feature fusion, the feature vector is denoted as $X\in \mathbb{R}^{F \times C \times T_1}$, where $F=5 \times B$ denotes the number of features, $C$ denotes the number of spatial channels, and $T_1$ denotes the number of temporal epochs. 

\subsection{Pre-trained data augmentation module}
\label{sec3.2}
To mitigate overfitting and training instability caused by the scarcity of dementia subtype samples, such as FTD and VCI, we employ a pretraining strategy based on a diffusion model. The model is first pre-trained on a large-scale independent dataset to capture general EEG generative patterns and subsequently fine-tuned on target datasets with a reduced learning rate. This strategy enables the model to adapt to subtype-specific data distributions while benefiting from general EEG generative representations.

As shown in Fig.~\ref{model}b, the diffusion model contains two processes: the forward diffusion process and the reverse generation process. The forward process $q$ gradually corrupts the data distribution $O_0 \sim P_{data}(X)$ by adding Gaussian noise over $D$ steps. This is formulated as a Markov chain:
\begin{equation} 
    q(O_d | O_{d-1}) = \mathcal{N}(O_d; \sqrt{1-\beta_d}O_{d-1}, \beta_d \mathbf{I}),
    \label{eq1}
\end{equation}
\begin{equation}
    q(O_{1:D}|O_0) = \prod_{d=1}^{D} q(O_d|O_{d-1}),
\end{equation}
where $\{\beta_d\}_{d=1}^D \in (0,1)$ represents the noise variance schedule, and $\mathbf{I}$ denotes the identity matrix. 

The reverse process aims to recover the original data $O_0$ from pure Gaussian noise $O_D \sim \mathcal{N}(0, \mathbf{I})$. This denoising process is modeled by a neural network parameterized by $\theta$, defined as:
\begin{equation}
    p_\theta(O_{d-1}|O_d) = \mathcal{N}(O_{d-1}; \mu_\theta(O_d, d), \Sigma_\theta(O_d, d)),
\end{equation}
\begin{equation}
    p_\theta(O_{0:{D-1}}|O_D) = \prod_{d=1}^{D} p_\theta(O_{d-1}|O_d),
\end{equation}
where $\mu_\theta$ and $\Sigma_\theta$ predict the mean and variance of the posterior distribution, respectively. 

The generated EEG features are mixed with the real EEG features, formulated as $O_D \sim \mathcal{N}(0, \mathbf{I}), X \leftarrow X \cup \{\text{PTDA}(O_D)\}$, to serve as the augmented input for subsequent processing.

\subsection{Gated spatiotemporal attention module}
EEG exhibits complex spatiotemporal dependencies across brain regions and temporal domains, where both local neural patterns and long-range interactions are critical for dementia analysis.
To effectively capture long-range spatial dependencies and temporal dynamics from EEG, we propose the GSA module. 

As illustrated in Fig.~\ref{model}c, the GSA module comprises two primary sub-modules: the gated Spatiotemporal Attention Unit (gSTA) and the Convolutional Feedforward Neural Network (ConvFFN) \cite{xie2021segformer}.
For the $k$-th GSA block $(k \in \{1, \dots, K\})$, the process from input $X_{k-1}$ to output $X_k$ is formulated as follows, where $X_0=X$:
\begin{equation}
    \begin{aligned}
    X_{k}^\mathrm{G}=\mathrm{gSTA}\bigl(\mathrm{BN}({X}_{k-1})\bigr),\\
    {X}_{k}' = {X}_{k-1} + \mathrm{DropPath}\Bigl(\lambda_1 \odot X_{k}^\mathrm{G}\Bigr), 
    \end{aligned}
\end{equation}
\begin{equation}
    \begin{aligned}
    {X}_{k}^\mathrm{F} = \mathrm{ConvFFN}\bigl(\mathrm{BN}({X}'_{k})\bigr),\\
    {X}_{k} = {X}_{k}' + \mathrm{DropPath}\Bigl(\lambda_2 \odot {X}_{k}^\mathrm{F}\Bigr), 
    \end{aligned}
\end{equation}
where $\lambda_1$ and $\lambda_2$ are learnable scaling parameters initialized to small values to stabilize the training of deep networks, and $\mathrm{DropPath} (\cdot)$ denotes the stochastic depth regularization used to prevent overfitting.

The gSTA unit models contextual information via a gated convolutional structure.
The input is first projected to a latent space, followed by local and comprehensive feature extraction:
\begin{equation}
    X_{k}^{\mathrm{L}} = \mathrm{DWConv}\bigl(\mathrm{Proj}\bigl(\mathrm{BN}({X}_{k-1})\bigr)\bigr),
\end{equation}
\begin{equation}    
    X_{k}^{\mathrm{C}} = \mathrm{DWDConv}(X_{k}^{\mathrm{L}},d=3),
\end{equation}
where $\mathrm{Proj}(\cdot)$ denotes a $1\times1$ convolution with GELU activation, $\mathrm{DWConv}(\cdot)$ denotes a depthwise convolution for local feature extraction, and $\mathrm{DWDConv}(\cdot,d=3)$ denotes a depthwise dilated convolution with a dilation factor of $3$ to expand the receptive field for modeling long-range dependencies.

The features are then split and modulated by a gating mechanism, as follows:
\begin{equation}
    [{X}_{k}^\mathrm{G1},{X}_{k}^\mathrm{G2}] = \mathrm{Split}\!\bigl(\mathrm{Conv}_{1\times1}({X}_{k}^{\mathrm{C}})\bigr),
\end{equation}
\begin{equation}
    X_{k}^\mathrm{G} = X_{k}^\mathrm{G1} \odot \sigma(X_{k}^\mathrm{G2}),
\end{equation}
where $\sigma(\cdot)$ denotes the sigmoid activation function.

To enhance channel-wise feature interaction and further refine spatiotemporal representations, we incorporate a ConvFFN \cite{xie2021segformer}. The specific computational process is defined as follows:
\begin{equation}
    {X_{k}^\mathrm{F}} = \mathrm{Conv}_{1\times1}\Bigl(\mathrm{GELU}\bigl(\mathrm{DWConv}\bigl(\mathrm{Conv}_{1\times1}(\mathrm{BN}({X}'_{k}))\bigr)\bigr)\Bigr),
\end{equation}

\subsection{Task-guided query module}
\label{sec3.4}
To bridge the semantic gap between shared spatiotemporal representations and distinct downstream tasks, we design the TGQ module for task-specific feature extraction.
Specifically, the spatiotemporal feature map ${X}_{K}$ is flattened and permuted into a sequence ${X}^\mathrm{{S}} \in \mathbb{R}^{(C \times T_1)\times E}$, where $E$ denotes the embedding dimension of the spatiotemporal features. We then initialize learnable task queries $Q_j \in \mathbb{R}^{E}$, where $(j \in \{\text{Dementia}, \text{MMSE}\})$ to interact with $X^\mathrm{S}$ via Multi-Head Cross-Attention (MHCA) \cite{vaswani2017attention}. 
The computation can be expressed as follows:
\begin{equation}
    {Q}_j',\; {K}',\; {V}' = \mathrm{LN}({Q}_{j}),\;\mathrm{LN}({X}^{\mathrm{S}}),\;\mathrm{LN}({X}^{\mathrm{S}}),
\end{equation}
\begin{equation}
    {X}_j^{\mathrm{A}} = \mathrm{MHCA}({Q_j}',{K}',{V}') + {Q}_{{j}},
\end{equation}
where $\mathrm{LN}(\cdot)$ denotes layer normalization and $\mathrm{MHCA}(\cdot)$ utilizes scaled dot-product attention to measure the relevance between task queries and sequences.
The task-guided representation ${X}_{j}^\mathrm{A}$ is refined via a Feed-forward Network (FFN) with residual connections:
\begin{equation}
X_j^{\mathrm{Q}} = {X}_{j}^\mathrm{A} + \mathrm{DropPath}(\mathrm{FFN}({X}_{j}^\mathrm{A})),
\end{equation}

To balance convergence speed and both task objectives during joint training, we employ a dynamic weighting strategy for collaborative optimization.
We calculate the loss ratios $w_e^{\rm CE}$ and $w_e^{\rm MSE}$ to determine the average weights $\lambda _e^{\rm CE}$ and $\lambda _e^{\rm MSE}$ for the two tasks. 
\begin{equation}
    {w_e^{\rm CE} } = \frac{\mathcal{L}_e^{\rm CE} } {\mathcal{L}_{e-1}^{\rm CE} }, 
    {w_e^{\rm MSE} } = \frac{\mathcal{L}_e^{\rm MSE} } {\mathcal{L}_{e-1}^{\rm MSE} },
\end{equation}
\begin{equation}
\lambda _e^{\rm CE} = \frac{  N\exp \left( {{w_e^{\rm CE} } / {\xi} }\right) }{\exp \left( {{w_e^{\rm CE} } / {\xi} } \right)  +\exp \left( {{w_e^{\rm MSE} } / {\xi} } \right) },
	\label{eq33}
\end{equation}
\begin{equation}
	\lambda _e^{\rm MSE} =\frac{ N\exp \left({{w_e^{\rm MSE} } / {\xi} } \right) }{\exp \left({{w_e^{\rm CE} } / {\xi } } \right) + \exp \left({{w_e^{\rm MSE} } / {\xi } } \right) },
	\label{eq34}
\end{equation}
where $\mathcal{L}_e^{\rm CE}$ and $\mathcal{L}_e^{\rm MSE}$ respectively represent the Cross-Entropy (CE) and Mean Squared Error (MSE) losses for dementia diagnosis and MMSE prediction at epoch $e$, $\xi$ denotes a temperature parameter used to balance the optimization rates of the two tasks, and $N$ denotes the task number. For the initial epoch ($e=1$), since the previous loss $\mathcal{L}_{0}$ is undefined, the task weights are initialized equally as $\lambda_1^{\text{CE}} = \lambda_1^{\text{MSE}} = 1$. It is worth noting that the sum of the weights is fixed, such that $\lambda _e^{\rm CE} + \lambda _e^{\rm MSE}=N$.

The total loss $\mathcal{L}_e$ at the $e$-th epoch is defined as follows:
\begin{equation}
	\mathcal{L}_e = \lambda _e^{\rm CE} \mathcal{L}_e^{\rm CE} + \lambda _e^{\rm MSE} \mathcal{L}_e^{\rm MSE}.
	\label{eq35}
\end{equation}

\begin{table}
	\centering
	\caption{\textbf{Demographic information for datasets.}}\label{dataset}
    \vspace{-5pt}
		\begin{tabular}{@{}ccccc@{}}
			\toprule
			{\textbf{Dataset}}&
            \textbf{Type}&{\textbf{Female/male}} & {\textbf{Age}} & {\textbf{MMSE}} \\
              \midrule
            \multirow{3}{*}{\textbf{XY02}} &
			  AD & 42/14 & 67.50$\pm$11.00  & 10.57$\pm$7.42  \\ 
			  & FTD & 12/16 & 63.14$\pm$9.31  & 12.33$\pm$7.23  \\ 
			  & VCI & 15/16 & 67.94$\pm$9.50  & 14.57$\pm$6.40  \\
			\midrule
            \multirow{3}{*}{\textbf{DS004504}} &
            AD & 24/12 & 66.40$\pm$7.90  & 17.75$\pm$4.44 \\ 
			& FTD & 9/14 & 63.60$\pm$8.20  & 22.17$\pm$2.58  \\
            & CN & 11/18 & 67.90$\pm$5.40 & 30.00$\pm$0.00 \\
            
			\bottomrule
	\end{tabular}
    \vspace{-10pt}
\end{table}

\section{Experiment}
\subsection{Dataset}

Three independent datasets were utilized in this study, with demographics detailed in Table~\ref{dataset}. 
\begin{itemize}
    \item XY01 includes 485 cognitively normal (CN) participants, while XY02 comprises 56 AD, 31 VCI, and 28 FTD participants. Both datasets were collected at Xiangya Hospital. Continuous EEG was recorded for approximately 8–15 minutes while participants remained in a resting state with eyes closed, under the supervision of a physician. EEG recordings were acquired using a Nicolet EEG system (V32-20020130) at a sampling rate of 200 Hz, with 21 electrodes positioned according to the international 10–20 system. Among them, earlobe electrodes (A1, A2) were used as reference electrodes for positioning standard scalp electrodes including Fp1, Fp2, F3, F4, C3, C4, P3, P4, F7, F8, T3, T4, T5, T6, Fz, Cz, Pz, A1, A2, O1, and O2. The unlabeled XY01 dataset was used for pretraining the diffusion model, while XY02 served as the primary dataset for fine-tuning the network. 
    \item DS004504 \cite{miltiadous2023dataset} (36 AD, 23 FTD, 29 CN samples) was used for external validation to assess the model's cross-dataset generalization. It was recorded using a Nihon Kohden EEG 2100 system with 19 electrodes in an eyes-closed sitting position. 
\end{itemize}

Significant differences in MMSE among diagnostic groups (Kruskal-Wallis test, $p < 0.05$) reveal a strong correlation between the dementia diagnosis and MMSE prediction tasks. All ethical and experimental procedures were approved by the Human Ethics Committee of Central South University and conducted in accordance with the Declaration of Helsinki, with written informed consent obtained from all subjects.

The raw EEG was preprocessed using the \texttt{Python-MNE} toolbox:
\begin{itemize}
    \item \textbf{Electrode Localization:}  
    Channels were spatially aligned to the standard international 10-20 system.
    \item \textbf{Filtering:}  
    A 1--55~Hz band-pass filter and a 50~Hz notch filter were applied to suppress noise and power-line interference.
    \item \textbf{Artifact Removal:}  
    Independent Component Analysis (ICA) was utilized to identify and reject ocular and muscular artifacts.
    \item \textbf{Frequency Band Decomposition:} 
    Five frequency bands—Delta (1–4 Hz), Theta (4–8 Hz), Alpha (8–13 Hz), Beta (13–30 Hz), and Gamma (30–50 Hz)—were extracted from the EEG using band-pass filtering.
    \item \textbf{Epoch Segmentation:} Each band was segmented into non-overlapping 2s epochs.

\end{itemize}

\begin{table*}[!t]
\vspace{-5mm}
	\centering
	\caption{\textbf{Performance comparison of TGSN with SOTA methods on the XY02 dataset (mean $\pm$ standard deviation). The best results are highlighted in bold.}}\label{comparison}
     \vspace{-5pt}
     \renewcommand\arraystretch{0.85}
     \resizebox{\textwidth}{!}{
			\begin{tabular}{ccccccc}
			\toprule
            \multirow{2}{*}{\textbf{Task}} & \multirow{2}{*}{\textbf{Method}} & \multicolumn{4}{c}{\textbf{Diagnosis}} & \multicolumn{1}{c}{\textbf{MMSE}}  \\
            \cmidrule(lr){3-6} \cmidrule(lr){7-7}
		      & & \textbf{ACC (\%) $\uparrow$} & \textbf{AUC (\%) $\uparrow$}  & \textbf{SEN (\%) $\uparrow$} & \textbf{SPE (\%) $\uparrow$} & \textbf{RMSE $\downarrow$} \\ 
            \midrule
                \multirow{10}{*}{AD/FTD} 
                & MNet \cite{watanabe2024deep} & 81.27$\pm$3.96 & 74.24$\pm$6.00 & 53.20$\pm$12.27 & 95.28$\pm$1.38    &- \\
                & SpectroCVT-Net \cite{bravo2024spectrocvt} & 81.39$\pm$5.61 & 77.78$\pm$5.94 & 70.82$\pm$3.97 & 84.73$\pm$1.96      &- \\
                & SATNet \cite{sibilano2024understanding} & 77.24$\pm$1.03  & 74.20$\pm$0.99 & 63.60$\pm$2.59 & 84.80$\pm$2.11    &- \\
                \cmidrule(lr){2-7}
                & MLBC \cite{lin2021predicting} & - & - & - & - & 4.04$\pm$0.17  \\
                & RankCNN \cite{qiao2022ranking} & - & - & - & - & 3.85$\pm$0.15 \\
                &AdaSVM \cite{sun2024ensemble} & - & - & - & - & 3.37$\pm$0.18 \\
                \cmidrule(lr){2-7}
                & GMSS \cite{li2022gmss} & 78.55$\pm$4.44 & 77.50$\pm$6.38 & 69.00$\pm$8.35 & 86.00$\pm$7.12 & 4.03$\pm$0.69 \\
                & CAW-MAS-STST \cite{luo2023dual} & 79.93$\pm$4.16 & 73.09$\pm$4.70 & 51.33$\pm$7.06 & 94.85$\pm$2.75 & 4.92$\pm$0.63 \\
                & GPFMTL \cite{wang2025gated} & 76.27$\pm$4.26 & 85.88$\pm$2.82 & 74.04$\pm$4.55 & 82.69$\pm$3.03 & 3.53$\pm$0.28 \\
                & \textbf{TGSN}  & \textbf{97.78$\pm$1.59} & \textbf{97.40$\pm$1.86} & \textbf{98.93$\pm$2.86} & \textbf{95.00$\pm$3.57} & \textbf{1.93$\pm$0.27} \\
            \midrule
            \multirow{10}{*}{AD/VCI} 
                & MNet \cite{watanabe2024deep} & 81.37$\pm$1.21 & 75.23$\pm$1.92 & 53.95$\pm$4.64 & \textbf{96.55$\pm$1.04} &- \\
                & SpectroCVT-Net \cite{bravo2024spectrocvt} & 83.98$\pm$0.57 & 85.67$\pm$0.58 & 91.57$\pm$0.56 & 79.77$\pm$0.60  &- \\
                & SATNet \cite{sibilano2024understanding} & 80.44$\pm$4.00  & 74.36$\pm$3.38 & 62.49$\pm$2.42 & 86.22$\pm$4.34 &- \\
                \cmidrule(lr){2-7}
                & MLBC \cite{lin2021predicting} & - & - & - & - & 4.65$\pm$0.61  \\
                & RankCNN \cite{qiao2022ranking} & - & - & - & - & 4.29$\pm$1.85 \\
                &AdaSVM \cite{sun2024ensemble} & - & - & - & - & 3.62$\pm$0.34 \\
                \cmidrule(lr){2-7}
                & GMSS \cite{li2022gmss} & 76.32$\pm$5.11 & 75.58$\pm$6.34 & 72.67$\pm$9.71 & 78.48$\pm$3.44  & 5.56$\pm$1.83 \\
                & CAW-MAS-STST \cite{luo2023dual} & 81.10$\pm$3.03 & 75.09$\pm$5.10 & 55.33$\pm$8.79 & 94.85$\pm$3.75 & 4.98$\pm$0.59 \\
                & GPFMTL \cite{wang2025gated} & 71.26$\pm$3.35 & 81.28$\pm$2.91 & 73.72$\pm$3.89 & 82.48$\pm$2.60  & 4.69$\pm$0.72 \\
                & \textbf{TGSN}  & \textbf{94.59$\pm$0.70} & \textbf{94.60$\pm$0.71} & \textbf{98.57$\pm$1.25} & 90.63$\pm$0.26 & \textbf{2.13$\pm$0.51} \\
            \midrule
            \multirow{10}{*}{FTD/VCI} 
                & MNet \cite{watanabe2024deep} & 78.98$\pm$2.45 & 78.90$\pm$2.55 & 79.13$\pm$3.19 & 78.67$\pm$2.19 &- \\
                & SpectroCVT-Net \cite{bravo2024spectrocvt} & 80.22$\pm$1.82 & 79.78$\pm$2.27 & 78.50$\pm$4.67 & 81.07$\pm$2.50 &- \\
                & SATNet \cite{sibilano2024understanding} & 77.67$\pm$1.25  & 74.50$\pm$1.82 & 64.67$\pm$6.86 & 84.33$\pm$4.13  &- \\
                \cmidrule(lr){2-7}
                & MLBC \cite{lin2021predicting} & - & - & - & - & 4.70$\pm$1.30  \\
                & RankCNN \cite{qiao2022ranking} & - & - & - & - & 4.31$\pm$0.69 \\
                &AdaSVM \cite{sun2024ensemble} & - & - & - & - & 4.32$\pm$0.52 \\
                \cmidrule(lr){2-7}
                & GMSS \cite{li2022gmss} & 73.85$\pm$5.07 & 74.25$\pm$4.32 & 72.50$\pm$12.17 & 76.00$\pm$7.80 & 4.75$\pm$1.47 \\
                & CAW-MAS-STST \cite{luo2023dual} & 76.78$\pm$2.03 & 73.30$\pm$6.61 & 62.67$\pm$10.84 & 83.93$\pm$5.75 & 4.73$\pm$0.81 \\
                & GPFMTL \cite{wang2025gated} & 84.19$\pm$1.54 & 85.43$\pm$4.88 & 83.44$\pm$0.87 & 88.96$\pm$0.58  & 5.18$\pm$1.03 \\
                & \textbf{TGSN}  & \textbf{92.28$\pm$2.92} & \textbf{92.30$\pm$2.92} & \textbf{89.38$\pm$5.27} & \textbf{95.22$\pm$4.52} & \textbf{2.95$\pm$0.23}  \\
            \midrule
            \multirow{10}{*}{AD/FTD/VCI} 
                & MNet \cite{watanabe2024deep} & 74.49$\pm$2.17 & 83.74$\pm$4.74 & 66.12$\pm$2.04 & 83.76$\pm$1.08 &- \\
                & SpectroCVT-Net \cite{bravo2024spectrocvt} & 69.86$\pm$4.88 & 82.94$\pm$5.18 & 69.01$\pm$4.52 & 83.88$\pm$2.32 &- \\
                & SATNet \cite{sibilano2024understanding} & 75.65$\pm$2.13  & 85.13$\pm$4.91 & 66.67$\pm$1.86 & 84.33$\pm$4.13  &- \\
                \cmidrule(lr){2-7}
                & MLBC \cite{lin2021predicting} & - & - & - & - & 3.81$\pm$0.46  \\
                & RankCNN \cite{qiao2022ranking} & - & - & - & - & 4.38$\pm$0.93 \\
                &AdaSVM \cite{sun2024ensemble} & - & - & - & - & 4.10$\pm$0.24 \\
                \cmidrule(lr){2-7}
                & GMSS \cite{li2022gmss} & 71.88$\pm$5.91 & 79.73$\pm$3.20 & 67.72$\pm$3.73 &    83.62$\pm$2.52 & 4.62$\pm$0.67 \\
                & CAW-MAS-STST \cite{luo2023dual} & 73.04$\pm$3.25 & 84.12$\pm$5.08 & 66.54$\pm$3.89 & 83.56$\pm$1.93 & 4.50$\pm$0.35 \\
                & GPFMTL \cite{wang2025gated} & 69.57$\pm$2.46 & 80.23$\pm$1.56 & 65.58$\pm$2.80 &    82.47$\pm$1.39 & 4.62$\pm$0.66 \\
                & \textbf{TGSN}  & \textbf{83.93$\pm$1.17} & \textbf{87.95$\pm$0.90} & \textbf{83.95$\pm$1.20} & \textbf{91.96$\pm$0.60} & \textbf{2.38$\pm$0.25} \\
            \bottomrule
	       \end{tabular}}
\vspace{-5mm}
\end{table*}

\begin{table}[htbp]
\vspace{-1mm}
	\centering
	\caption{\textbf{Performance and comparison on the DS004504 dataset. Best results are highlighted in bold.}}\label{ds004504}
    \vspace{-5pt}
		\renewcommand{\arraystretch}{0.85}
        \setlength\tabcolsep{4pt} 
		\begin{tabular}{@{}ccccc@{}}
		\toprule
		\multirow{2}{*}{\textbf{Method}} & \multirow{2}{*}{\textbf{Task}} & \multicolumn{2}{c}{\textbf{Diagnosis}} & \textbf{MMSE} \\
		\cmidrule(lr){3-4} \cmidrule{5-5} 
		  & & \textbf{ACC (\%) $\uparrow$} & \textbf{AUC (\%) $\uparrow$} & \textbf{RMSE $\downarrow$}\\ 
		\midrule
        \multirow{4}{*}{\shortstack{Chen \\et al. \cite{chen2023multi}}}
        & {AD/FTD} & 81.56 & 82.19 & - \\
        & {AD/CN} & 87.33 & 88.19 & - \\
        & {FTD/CN} & 82.98 & 83.41 & - \\
        & {AD/FTD/CN} & 79.12 & 80.23 & - \\
        \midrule
        \multirow{3}{*}{\shortstack{Si et al.\\ \cite{si2023differentiating}}}
        & {AD/FTD} & 81.10 & 90.10 & - \\
        & {AD/CN} & 86.20 & 90.70 & - \\
        & {FTD/CN} & 84.70 & 90.70 & - \\
        \midrule
        \multirow{3}{*}{\shortstack{Zheng \\et al. \cite{zheng2024multi}}}
        & {AD/FTD} & 72.88 & - & - \\
        & {AD/CN} & 87.69 & - & - \\
        & {FTD/CN} & 82.69 & - & - \\
        & {AD\&FTD/CN} & 86.36 & - & - \\
        \midrule
        \multirow{4}{*}{\shortstack{Kachare \\et al. \cite{kachare2024steadynet}}}
         & {AD/FTD} & 89.40 & 90.13 & - \\
         & {AD/CN} & 88.00 & 87.18 & - \\
         & {FTD/CN} & 92.25 & 90.35 & - \\
         & {AD/FTD/CN} & 84.59 &  88.02 & - \\
         \midrule
         \multirow{1}{*}{\shortstack{Taub et al. \cite{taub2024ranking}}}
        & AD/CN & 81.34$\pm$0.60 & - & - \\
        \midrule
        \multirow{4}{*}{\shortstack{Lee \\et al. \cite{lee2025diagnosis}}}
        & AD/FTD & - & - & 3.91$\pm$0.02 \\
        & AD/CN & 88.60 & 88.30 & - \\
        & FTD/CN & \textbf{96.00} & 95.00 & - \\
        & AD\&FTD/CN & 92.50 & 90.40 & - \\
        \midrule
        \multirow{4}{*}{\shortstack{Stefanou \\et al. \cite{stefanou2025novel}}}
        & AD/CN & 79.45$\pm$7.06 & - & - \\
        & FTD/CN & 72.85$\pm$3.08 & - & - \\
        & AD\&FTD/CN & 80.69$\pm$2.25 & - & - \\
        & AD/FTD/CN & 54.28$\pm$5.92 & - & - \\
        \midrule
        \multirow{3}{*}{\shortstack{Kerchove \\et al. \cite{kerchove2025decoding}}}
        & AD/FTD & 74.00 & - & - \\
        & AD/CN & 85.00 & - & - \\
        & FTD/CN & 74.00 & - & - \\
        \midrule
        \multirow{1}{*}{\shortstack{Cai et al. \cite{cai2026multi}}}
        & AD/FTD & 81.36 & 86.60 & - \\
        \midrule
        \multirow{4}{*}{TGSN}
         & {AD/FTD} & \textbf{93.08$\pm$1.12} & \textbf{98.69$\pm$0.13} & \textbf{2.64$\pm$0.38} \\
         & {AD/CN} & \textbf{92.01$\pm$0.57} & \textbf{98.95$\pm$0.21} & \textbf{5.33$\pm$0.75} \\
         & {FTD/CN} & 93.35$\pm$0.75 & \textbf{99.15$\pm$0.18} & \textbf{4.42$\pm$0.84} \\
         & {AD/FTD/CN} & \textbf{89.83$\pm$1.12} &  \textbf{92.39$\pm$0.85} & \textbf{4.15$\pm$0.36} \\
        \bottomrule
		\end{tabular}
    \vspace{-5mm}
\end{table}
\begin{table*}[!htbp]
\vspace{-5mm}
	\centering
	\caption{\textbf{Ablation study on the contribution of individual EEG bands (mean $\pm$ standard deviation). Best results are highlighted in bold.}}\label{bands}
    \vspace{-5pt}
    \renewcommand\arraystretch{0.85}
	\resizebox{\textwidth}{!}{
	\begin{tabular}{ccccccccccc}
		\toprule
         \multirow{2}{*}{\textbf{Task}} & 
         \multirow{2}{*}{\textbf{Delta}} &  \multirow{2}{*}{\textbf{Theta}} & \multirow{2}{*}{\textbf{Alpha}} & 
         \multirow{2}{*}{\textbf{Beta}} & 
         \multirow{2}{*}{\textbf{Gamma}} &
         \multicolumn{4}{c}{\textbf{Diagnosis}} & \multicolumn{1}{c}{\textbf{MMSE}}  \\
          \cmidrule(lr){7-10} \cmidrule(lr){11-11}
		& & & & & & \textbf{ACC (\%) $\uparrow$} & \textbf{AUC (\%) $\uparrow$}  & \textbf{SEN (\%) $\uparrow$} & \textbf{SPE (\%) $\uparrow$} & \textbf{RMSE $\downarrow$} \\ 
		\midrule
		\multirow{9}{*}{AD/FTD} 
		& $\checkmark$ & - & - & - & - & 83.78$\pm$1.21 & 83.46$\pm$1.63 & 85.68$\pm$1.66 & 81.25$\pm$4.68 & 1.71$\pm$0.15 \\
		& - & $\checkmark$ & - & - & - & 79.70$\pm$1.45 & 80.03$\pm$1.57 & 77.97$\pm$2.23 & 82.08$\pm$3.47 & 2.16$\pm$0.14 \\
		& - & - & $\checkmark$ & - & - & 83.07$\pm$1.01 & 83.23$\pm$1.02 & 82.29$\pm$0.94 & 84.17$\pm$1.28 & 2.02$\pm$0.13 \\
		& - & - & - & $\checkmark$ & - & 84.66$\pm$1.84 & 84.00$\pm$1.55 & 88.62$\pm$3.92 & 79.37$\pm$2.22 & 1.98$\pm$0.16 \\
        & - & - & - & - & $\checkmark$ & 83.87$\pm$1.27 & 83.10$\pm$1.01 & 88.27$\pm$3.16 & 77.92$\pm$1.93 & 2.14$\pm$0.14 \\
		& - & - & - & $\checkmark$ & $\checkmark$ & 90.43$\pm$1.97 & 89.71$\pm$1.70 & 94.63$\pm$2.52 & 84.79$\pm$1.98 & 1.56$\pm$0.12 \\
		& $\checkmark$ & - & - & $\checkmark$ & $\checkmark$ & 94.01$\pm$2.27 & 94.19$\pm$2.51 & 93.18$\pm$3.32 & \textbf{95.21$\pm$0.29} & \textbf{1.51$\pm$0.14} \\
		& $\checkmark$ & - & $\checkmark$ & $\checkmark$ & $\checkmark$ & 97.70$\pm$1.64 & 97.35$\pm$1.87 & 98.70$\pm$0.77 & 95.00$\pm$1.02 & 1.55$\pm$0.11 \\
        \cmidrule{7-11}
        & $\checkmark$ & $\checkmark$ & $\checkmark$ & $\checkmark$ & $\checkmark$ & \textbf{97.78$\pm$1.59} & \textbf{97.40$\pm$1.86} & \textbf{98.93$\pm$2.86} & 95.00$\pm$3.57 & 1.93$\pm$0.27 \\
		\midrule
		\multirow{9}{*}{AD/VCI} 
	    & $\checkmark$ & - & - & - & - & 80.69$\pm$1.30 & 80.67$\pm$1.31 & 89.51$\pm$1.65 & 71.83$\pm$4.02 & 4.12$\pm$0.17 \\
		& - & $\checkmark$ & - & - & - & 81.56$\pm$1.32 & 81.56$\pm$1.34 & 80.70$\pm$2.33 & 82.42$\pm$1.90 & 3.84$\pm$0.06 \\
		& - & - & $\checkmark$ & - & - & 84.04$\pm$1.77 & 84.04$\pm$1.80 & 81.42$\pm$1.89 & 86.67$\pm$3.40 & 3.55$\pm$0.07 \\
		& - & - & - & $\checkmark$ & - & 82.63$\pm$1.01 & 82.64$\pm$1.05 & 84.27$\pm$2.79 & 81.01$\pm$4.87 & 3.35$\pm$0.05 \\
        & - & - & - & - & $\checkmark$ & 80.67$\pm$1.38 & 80.70$\pm$1.38 & 79.49$\pm$0.64 & 81.91$\pm$5.28 & 3.76$\pm$0.51 \\
		& - & - & $\checkmark$ & $\checkmark$ & - & 87.49$\pm$2.29 & 87.51$\pm$2.33 & 86.73$\pm$2.40 & 88.29$\pm$7.05 & \textbf{2.80$\pm$0.48} \\
		& - & $\checkmark$ & $\checkmark$ & $\checkmark$ & - & 90.96$\pm$2.75 & 90.97$\pm$2.70 & 84.41$\pm$5.88 & \textbf{97.52$\pm$3.14} & 2.94$\pm$0.64 \\
		& $\checkmark$ & $\checkmark$ & $\checkmark$ & $\checkmark$ & - & 93.35$\pm$0.44 & 93.37$\pm$0.42 & 93.46$\pm$1.33 & 93.29$\pm$1.32 & 3.22$\pm$0.55 \\
        \cmidrule{7-11}
        & $\checkmark$ & $\checkmark$ & $\checkmark$ & $\checkmark$ & $\checkmark$ & \textbf{94.51$\pm$0.33} &\textbf{ 94.52$\pm$0.33} & \textbf{98.57$\pm$1.34} & 90.46$\pm$1.15 & 2.99$\pm$0.31 \\
		\midrule
		\multirow{9}{*}{FTD/VCI}
	    & $\checkmark$ & - & - & - & - & 74.21$\pm$1.31 & 74.21$\pm$1.32 & 73.09$\pm$2.22 & 75.33$\pm$3.95 & 2.76$\pm$0.43 \\
		& - & $\checkmark$ & - & - & - & 66.76$\pm$0.98 & 66.74$\pm$1.02 & 66.15$\pm$2.91 & 67.34$\pm$4.57 & 3.09$\pm$0.47 \\
		& - & - & $\checkmark$ & - & - & 63.30$\pm$0.38 & 63.30$\pm$0.38 & 63.86$\pm$5.43 & 62.75$\pm$5.32 & 3.17$\pm$0.11 \\
		& - & - & - & $\checkmark$ & - & 68.62$\pm$0.44 & 68.61$\pm$0.44 & 74.84$\pm$3.60 & 62.38$\pm$3.09 & 2.85$\pm$0.39 \\
        & - & - & - & - & $\checkmark$ & 75.44$\pm$1.76 & 75.42$\pm$1.72 & 80.84$\pm$3.34 & 70.00$\pm$1.52 & 2.83$\pm$0.43 \\
		& - & - & - & $\checkmark$ & $\checkmark$ & 83.16$\pm$1.63 & 83.14$\pm$1.64 & 82.77$\pm$1.74 & 83.51$\pm$1.60 & 2.84$\pm$0.29 \\
		& $\checkmark$ & - & - & $\checkmark$ & $\checkmark$ & 85.38$\pm$1.52 & 85.35$\pm$1.53 & 84.73$\pm$2.52 & 85.97$\pm$2.09 & 2.93$\pm$0.34 \\
		& $\checkmark$ & - & $\checkmark$ & $\checkmark$ & $\checkmark$ & 88.64$\pm$3.49 & 88.65$\pm$3.51 & 86.88$\pm$3.67 & 90.43$\pm$4.88 & 2.96$\pm$0.32 \\
        \cmidrule{7-11}
        & $\checkmark$ & $\checkmark$ & $\checkmark$ & $\checkmark$ & $\checkmark$ & \textbf{92.28$\pm$2.92} & \textbf{92.30$\pm$2.92} & \textbf{89.38$\pm$5.27} & \textbf{95.22$\pm$4.52} & \textbf{2.95$\pm$0.23} \\
		\midrule
        \multirow{9}{*}{AD/FTD/VCI} 
		& $\checkmark$ & - & - & - & - & 64.12$\pm$1.10 & 73.09$\pm$0.83 & 64.12$\pm$1.10 & 82.05$\pm$0.56 & 3.81$\pm$0.21 \\
		& - & $\checkmark$ & - & - & - & 58.04$\pm$3.18 & 68.54$\pm$2.37 & 58.06$\pm$3.16 & 79.02$\pm$1.58 & 3.64$\pm$0.34 \\
		& - & - & $\checkmark$ & - & - & 59.40$\pm$2.40 & 69.56$\pm$1.75 & 59.41$\pm$2.33 & 79.71$\pm$1.17 & 3.44$\pm$0.30 \\
		& - & - & - & $\checkmark$ & - & 64.30$\pm$1.04 & 73.20$\pm$0.80 & 64.27$\pm$1.07 & 82.14$\pm$0.53 & 3.27$\pm$0.21 \\
        & - & - & - & - & $\checkmark$ & 64.71$\pm$0.75 & 73.54$\pm$0.57 & 64.73$\pm$0.76 & 82.36$\pm$0.37 & 3.57$\pm$0.21 \\
		& - & - & - & $\checkmark$ & $\checkmark$ & 78.13$\pm$0.65 & 83.62$\pm$0.47 & 78.18$\pm$0.62 & 89.07$\pm$0.32 & 2.50$\pm$0.18 \\
		& $\checkmark$ & - & - & $\checkmark$ & $\checkmark$ & 78.25$\pm$1.43 & 83.71$\pm$1.11 & 78.29$\pm$1.49 & 89.13$\pm$0.74 & 2.53$\pm$0.19 \\
		& $\checkmark$ & - & $\checkmark$ & $\checkmark$ & $\checkmark$ & 83.22$\pm$1.80 & 87.43$\pm$1.35 & 83.25$\pm$1.80 & 91.61$\pm$0.90 & 2.51$\pm$0.20 \\
        \cmidrule{7-11}
        & $\checkmark$ & $\checkmark$ & $\checkmark$ & $\checkmark$ & $\checkmark$ & \textbf{83.93$\pm$1.17} & \textbf{87.95$\pm$0.90} & \textbf{83.95$\pm$1.20} & \textbf{91.96$\pm$0.60} & \textbf{2.38$\pm$0.25} \\
		\bottomrule
	\end{tabular}}
    \vspace{-2mm}
\end{table*}

\begin{table*}[!h]
\vspace{-5mm}
	\centering
	\caption{\textbf{Ablation study of PTDA, GSA, and TGQ in TGSN (mean $\pm$ standard deviation). Best results are highlighted in bold.}}
	\label{Ablation}
    \vspace{-5pt}
    \renewcommand\arraystretch{0.85}
	\resizebox{\textwidth}{!}{
	\begin{tabular}{ccccccccc}
		\toprule
         \multirow{2}{*}{\textbf{Task}} & 
         \multirow{2}{*}{\textbf{PTDA}}  &  \multirow{2}{*}{\textbf{GSA}} & \multirow{2}{*}{\textbf{TGQ}} & \multicolumn{4}{c}{\textbf{Diagnosis}} & \multicolumn{1}{c}{\textbf{MMSE}}  \\
          \cmidrule(lr){5-8} \cmidrule(lr){9-9}
		& & & &\textbf{ACC (\%) $\uparrow$} & \textbf{AUC (\%) $\uparrow$}  & \textbf{SEN (\%) $\uparrow$} & \textbf{SPE (\%) $\uparrow$} & \textbf{RMSE $\downarrow$} \\ 
		\midrule
		\multirow{8}{*}{AD/FTD} 
		& - & - & - & 72.79$\pm$4.95 & 72.66$\pm$3.56 & 73.54$\pm$9.34 & 71.78$\pm$3.46 & 4.07$\pm$0.93 \\
		& $\checkmark$ & - & - & 86.42$\pm$0.96 & 87.80$\pm$0.96 & 78.51$\pm$1.98 & 97.08$\pm$2.06 & \textbf{1.50$\pm$0.46} \\
		& - & $\checkmark$ & - & 75.49$\pm$4.32 & 74.46$\pm$6.75 & 75.81$\pm$5.93 & 73.11$\pm$16.67 & 4.60$\pm$0.61 \\
		& - & - & $\checkmark$ & 82.97$\pm$2.51 & 80.82$\pm$1.69 & 87.42$\pm$3.84  & 74.22$\pm$1.13 & 3.00$\pm$0.48 \\
		& $\checkmark$ & $\checkmark$& - & 91.04$\pm$2.82 & 91.79$\pm$2.18 & 86.71$\pm$6.46 & \textbf{96.87$\pm$2.22} & 1.77$\pm$0.32 \\
		& $\checkmark$ & - &$\checkmark$ & 94.86$\pm$2.38 & 95.02$\pm$1.97 & 94.00 $\pm$9.32 & 96.04$\pm$0.78 & 1.79$\pm$0.53 \\
		& - & $\checkmark$ & $\checkmark$ & 87.33$\pm$1.07 & 83.49$\pm$0.64 & 95.20$\pm$1.75 & 71.78$\pm$0.83 & 3.30$\pm$0.28 \\
		\cmidrule{2-9}
            &$\checkmark$ & $\checkmark$ &$\checkmark$ & \textbf{97.78$\pm$1.59} & \textbf{97.40$\pm$1.86} & \textbf{98.93$\pm$2.86} & 95.00$\pm$3.57 & 1.93$\pm$0.27 \\
		\midrule
		\multirow{8}{*}{AD/VCI} 
	    & - & - & - & 72.20$\pm$4.12 & 74.30$\pm$4.64 & 82.38$\pm$5.88 & 66.21$\pm$4.08 & 5.38$\pm$0.71 \\
	    & $\checkmark$ & - & - & 81.83$\pm$7.56 & 81.84$\pm$7.63 & 84.88$\pm$6.47 & 78.80$\pm$9.57  & 3.87$\pm$0.26 \\
	    & - &$\checkmark$ & - & 76.19$\pm$5.54 & 77.91$\pm$5.60 & 83.81$\pm$9.44 & 72.02$\pm$7.78 & 5.51$\pm$0.75 \\
	    & - & - & $\checkmark$ & 81.00$\pm$3.78 & 80.75$\pm$3.89 & 79.68$\pm$4.27  & 81.82$\pm$3.52 & 4.56$\pm$0.67 \\
	    & $\checkmark$ & $\checkmark$& - & 91.67$\pm$0.55 & 91.68$\pm$0.54 & 88.14$\pm$2.50 & 95.23$\pm$1.97 & 2.33$\pm$0.40 \\
	    & $\checkmark$ & - & $\checkmark$ & 86.51$\pm$3.17 & 86.51$\pm$3.20 & 76.37$\pm$6.24 & \textbf{96.64$\pm$1.11} & 2.94$\pm$0.34 \\
	    & - & $\checkmark$ & $\checkmark$ & 84.03$\pm$0.96 & 86.57$\pm$0.48 & 95.56$\pm$1.57 & 77.58$\pm$2.35 & 4.71$\pm$0.65 \\
	    \cmidrule{2-9}
	    & $\checkmark$ & $\checkmark$ & $\checkmark$ & \textbf{94.59$\pm$0.70} & \textbf{94.60$\pm$0.71} & \textbf{98.57$\pm$1.25} & 90.63$\pm$0.26 & \textbf{2.13$\pm$0.51} \\
    	\midrule
		\multirow{8}{*}{FTD/VCI}
	    & - & - & - & 65.51$\pm$4.14 & 66.48$\pm$3.93 & 54.29$\pm$8.28 & 78.67$\pm$1.63 & 4.32$\pm$0.97 \\
	    & $\checkmark$ & - & - & 73.65$\pm$5.99 & 73.63$\pm$6.01 & 68.57$\pm$5.57 & 78.70$\pm$6.63  & 2.57$\pm$0.32 \\
	    & - &$\checkmark$ & - & 70.00$\pm$6.50 & 70.37$\pm$7.16 & 68.73$\pm$8.08 & 72.00$\pm$7.82 & 4.33$\pm$0.92 \\
	    & - & - & $\checkmark$ & 70.61$\pm$7.86 & 71.10$\pm$8.01 & 75.08$\pm$4.80  & 67.11$\pm$2.49 & 4.30$\pm$1.05 \\
	    & $\checkmark$ & $\checkmark$& - & 82.36$\pm$0.70 & 82.33$\pm$0.68 & 80.10$\pm$0.55 & 84.56$\pm$1.27 & 2.28$\pm$0.40 \\
	    & $\checkmark$ & - & $\checkmark$ & 91.40$\pm$1.20 & 91.42$\pm$1.20
 & \textbf{91.48$\pm$2.01} & 91.36$\pm$1.73 & \textbf{2.08$\pm$0.44} \\
	    & - & $\checkmark$ & $\checkmark$ & 87.53$\pm$7.95 & 87.97$\pm$7.05 & 84.60$\pm$9.79 & 91.33$\pm$6.60 & 4.29$\pm$0.98 \\
	    \cmidrule{2-9}
	    &$\checkmark$ & $\checkmark$ &$\checkmark$ &  \textbf{92.28$\pm$2.92} & \textbf{92.30$\pm$2.92} & 89.38$\pm$5.27 & \textbf{95.22$\pm$4.52} & 2.95$\pm$0.23 \\
        \midrule
        \multirow{8}{*}{AD/FTD/VCI} 
		& - & - & - & 61.45$\pm$5.33 & 72.04$\pm$2.49 & 63.67$\pm$3.43 & 80.40$\pm$1.56 & 4.88$\pm$0.48\\
		& $\checkmark$ & - & - & 68.15$\pm$4.14 & 76.10$\pm$3.11 & 68.13$\pm$4.15 & 84.06$\pm$2.08 & 3.31$\pm$0.20\\
		& - & $\checkmark$ & - & 64.93$\pm$4.73 & 75.61$\pm$3.07 & 69.03$\pm$3.82 & 82.20$\pm$2.32 & 4.44$\pm$0.39\\
		& - & - & $\checkmark$ & 72.17$\pm$5.64 & 76.58$\pm$4.56 & 69.34$\pm$5.93  & 83.82$\pm$3.20 & 4.11$\pm$0.47\\
		& $\checkmark$ & $\checkmark$& - & 77.65$\pm$2.14 & 83.24$\pm$1.62 & 77.67$\pm$2.16 & 88.82$\pm$1.07 & 2.44$\pm$0.21 \\
		& $\checkmark$ & - &$\checkmark$ & 75.95$\pm$4.28 & 81.97$\pm$3.21 & 75.98$\pm$4.27  & 87.96$\pm$2.15 & 2.53$\pm$0.48 \\
		& - & $\checkmark$ & $\checkmark$ & 73.04$\pm$3.09 & 78.33$\pm$1.92 & 71.76$\pm$2.51
 & 84.91$\pm$1.40 & 3.84$\pm$0.34 \\
		\cmidrule{2-9}
            &$\checkmark$ & $\checkmark$ &$\checkmark$ & \textbf{83.93$\pm$1.17} & \textbf{87.95$\pm$0.90} & \textbf{83.95$\pm$1.20} & \textbf{91.96$\pm$0.60} & \textbf{2.38$\pm$0.25} \\
		\bottomrule
	\end{tabular}}
\end{table*}

\subsection{Setting and evaluation}
The proposed framework was implemented in PyTorch on an NVIDIA GeForce RTX4090 GPU. The framework was trained using the Adam optimizer with a learning rate of $1 \times 10^{-5}$ for a maximum of 500 epochs. An early stopping strategy with a patience of 100 epochs was applied to mitigate overfitting. The key hyperparameters include $K=3$ GSA blocks, $N=2$ tasks, and the temperature parameter $\xi=5$.

Experiments are conducted using k-fold (k=5) cross-validation with three independent random seeds, partitioning the dataset into training, validation, and test sets.
Performance metrics included Accuracy (ACC), Area Under the Curve (AUC), Sensitivity (SEN), and Specificity (SPE) for dementia diagnosis, and Root Mean Square Error (RMSE) for MMSE prediction.

\subsection{Comparison with SOTA methods}
To comprehensively evaluate the effectiveness of TGSN, we conducted k-fold (k=5) cross-validation against state-of-the-art (SOTA) methods on XY02, with quantitative results summarized in Table~\ref{comparison}.

\textbf{Comparison with dementia diagnosis methods:} 
MNet \cite{watanabe2024deep} utilizes a deep CNN with multimodal inputs, while SpectroCVT-Net \cite{bravo2024spectrocvt} combines CNNs and Transformers to extract local and global features. SATNet \cite{sibilano2024understanding} introduces attention mechanisms to focus on key brain regions. TGSN demonstrated superior performance across all diagnostic tasks. As shown in Table \ref{comparison}, it achieves an accuracy of 83.93\% in the three-class AD/FTD/VCI task, outperforming the best baseline SATNet by 8.28\%. For the binary AD/FTD task, TGSN improves the accuracy by 16.39\% over SpectroCVT-Net. These improvements can be attributed to our multi-band feature fusion module. Unlike baselines that treat frequency bands uniformly, TGSN adaptively focuses on subtype-specific EEG bands while simultaneously modeling spatiotemporal dynamics, yielding more discriminative representations.
    
\textbf{Comparison with MMSE prediction methods:}
MLBC \cite{lin2021predicting} employs feature selection with a Random Forest, RankCNN \cite{qiao2022ranking} utilizes a 3D-CNN with ranking loss, and AdaSVM \cite{sun2024ensemble} fuses ensemble learning with SVMs. While these methods are optimized specifically for MMSE prediction, they ignore the intrinsic correlation between disease diagnosis and cognitive scores. By jointly modeling these tasks, TGSN significantly reduces prediction error. In the AD/FTD/VCI task, our framework achieves a RMSE of 2.38, outperforming MLBC with $3.81$ and AdaSVM with $4.10$. This validates that collaborative learning leverages shared neurophysiological patterns to enhance prediction robustness.
    
\textbf{Comparison with MTL methods:} 
GMSS \cite{li2022gmss} uses graph-based self-supervised learning, CAW-MAS-STST \cite{luo2023dual} incorporates spectral and spatial features, and GPFMTL \cite{wang2025gated} employs a gated sharing mechanism. 
While these methods enable cross-task information sharing, they generally rely on fixed sharing strategies. In contrast, TGSN utilizes the TGQ module for flexible, task-specific feature interaction, consistently outperforming these MTL methods. 
For instance, in the challenging AD/FTD/VCI task, TGSN achieves a substantial 10.89\% accuracy improvement and a 2.12 RMSE reduction compared to CAW-MAS-STST. This demonstrates that task-specific feature extraction significantly outperforms fixed sharing strategies when handling heterogeneous objectives. By utilizing the TGQ module, TGSN effectively mitigates feature entanglement and achieves a balanced optimization across both tasks.

\subsection{Generalization to public dataset}
The framework's robustness and generalization capabilities were further validated on the public DS004504 dataset. As shown in Table \ref{ds004504}, TGSN demonstrates highly competitive performance.

In binary classification, TGSN achieves an accuracy of 92.01\% and an AUC of 98.95\% for the AD/CN task, markedly outperforming the 88.6\% and 86.2\% accuracies reported by \cite{lee2025diagnosis} and \cite{si2023differentiating}, respectively. For the challenging AD/FTD/CN task, the network yields an accuracy of 89.83\% and an AUC of 92.39\%, maintaining a clear advantage over existing baseline methods.
Beyond dementia diagnosis, TGSN simultaneously performs MMSE prediction. Notably, for the AD/FTD task, TGSN yields a RMSE of 2.64, achieving a significantly lower prediction error compared to the 3.91 reported by \cite{lee2025diagnosis}. Furthermore, the framework maintains robust performance across other tasks, yielding a RMSE of 4.15 in the AD/FTD/CN task and 5.33 in the AD/CN task. This substantial reduction in error indicates a markedly improved capability in modeling the complex relationship between EEG features and cognitive assessment.

Overall, these results demonstrate that the proposed TGSN framework can effectively integrate multi-band spatiotemporal EEG features and dynamically balance multiple tasks, leading to improved performance in both dementia diagnosis and MMSE prediction.

\subsection{Ablation study of different EEG bands}
To evaluate the diagnostic contribution of specific frequency bands, we conducted a stepwise frequency band fusion experiment. The model was initially trained on individual bands to establish baselines. Subsequently, bands were cumulatively added to the model one by one, prioritized from highest to lowest based on their single-band performance, until all five bands were included.
The results are summarized in Table \ref{bands}.

Single-band performance exhibits significant spectral heterogeneity across tasks. For instance, high-frequency components dominated in the AD/FTD task, with Beta and Gamma bands achieving accuracies of 84.66\% and 83.87\%, respectively. Conversely, in the FTD/VCI task, the Delta band achieved a classification accuracy of 74.21\%, demonstrating discriminative power comparable to the Gamma band, which reached 75.44\%. The performance of the Delta band likely reflects its ability to capture pathological slowing associated with vascular lesions.
Crucially, relying on any single band proves insufficient for complex diagnosis. As demonstrated in challenging AD/FTD/VCI tasks, the best-performing Gamma band achieved an accuracy of only 64.71\%. In contrast, the fusion of all five bands propelled the accuracy to 83.93\%. Similarly, for MMSE prediction, multi-band integration reduced the RMSE from 3.27 to 2.38. These substantial improvements confirm that complementary diagnostic information is distributed across the spectrum, validating the effectiveness of our multi-band feature fusion module.

\subsection{Ablation study of different modules}
The contribution of each core component was assessed via ablation studies, where we selectively removed or replaced the PTDA, GSA, and TGQ modules, as summarized in Table~\ref{Ablation}.

Removing the PTDA module consistently led to noticeable performance degradation across all binary and multi-class diagnostic tasks. In particular, for the most challenging multi-class diagnostic task, the classification accuracy dropped from 83.93\% to 73.04\%, indicating that diffusion-based pretraining substantially enhances representation robustness under limited samples. 
Replacing the GSA module with a MLP resulted in reduced ACC and AUC values. Specifically, accuracy decreased by approximately 8\% in both multi-class and AD/VCI tasks. 
This confirms the importance of GSA in modeling both long-range spatial dependencies and temporal dynamics.
Substituting the TGQ module with a fixed-weight loss mechanism compromised joint optimization. For example, the classification accuracy dropped from 83.93\% to 77.65\%, and the RMSE increased from 2.38 to 2.44 in the AD/FTD/VCI task. These results indicate that task-guided weighting is essential for balancing heterogeneous task objectives.
\begin{figure*}[!htbp]
\centering
\includegraphics[width=0.9\textwidth]{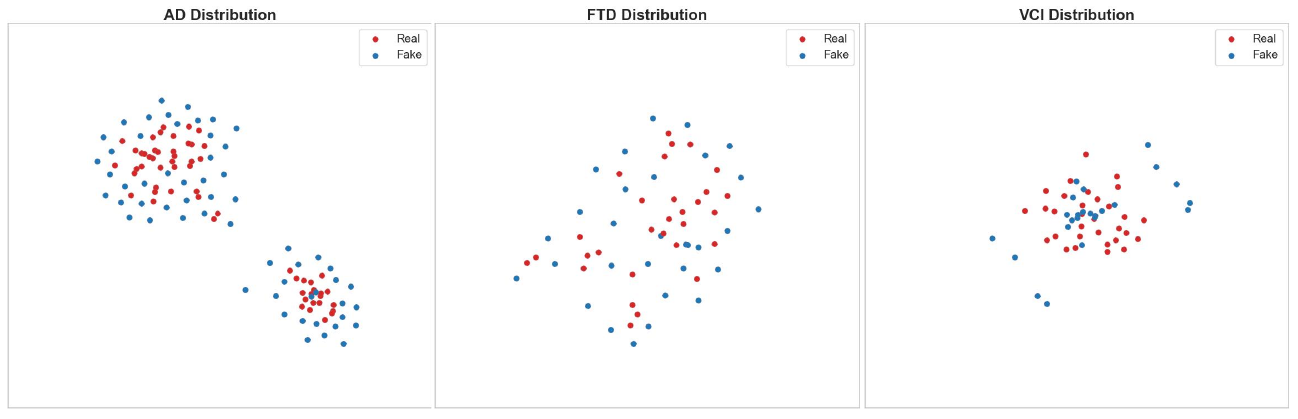}
\caption{\textbf{t-SNE visualization of feature distributions for real and generated EEG features across three classes.} } 
\label{tSNE}
\vspace{-2mm}
\end{figure*}
\begin{figure*}[!htbp]
\centering
\includegraphics[width=0.9\textwidth]{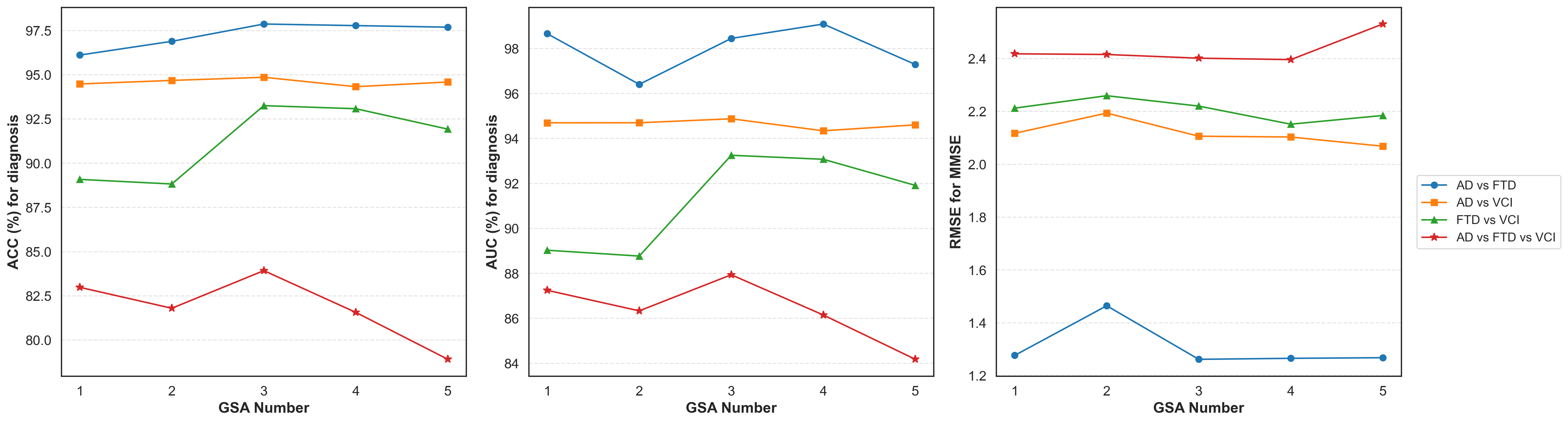}
\caption{\textbf{Sensitivity analysis of model performance with varying numbers of GSA layers.} } 
\label{GSA}
\vspace{-2mm}
\end{figure*}
\begin{figure}[!htbp]
\centering
\includegraphics[width=0.8\columnwidth]{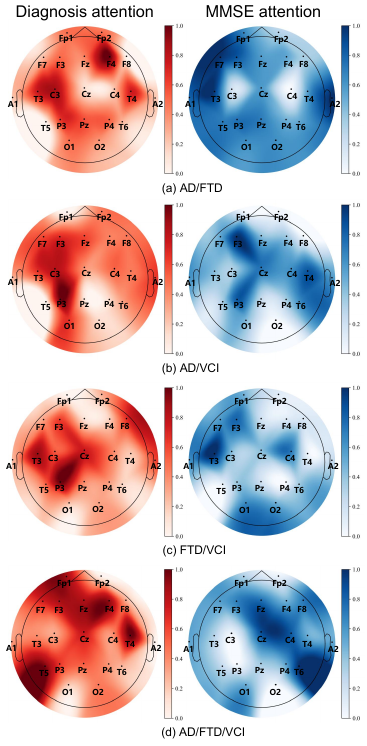}
\caption{
\textbf{Topographic maps of attention weights from the TGQ module.} 
Columns represent diagnosis and MMSE attention, while rows correspond to (a) AD/FTD, (b) AD/VCI, (c) FTD/VCI, and (d) AD/FTD/VCI tasks. 
The maps illustrate the spatial distribution of task-specific attention across EEG channels. 
Electrodes are grouped by anatomical regions: Fp1, Fp2, F3, F4, F7, F8, and Fz correspond to the frontal region; C3, C4, and Cz to the central region; P3, P4, and Pz to the parietal region; O1 and O2 to the occipital region; T3, T4, T5, and T6 to the temporal region; and A1 and A2 denote auxiliary electrodes.} 
\label{task}
\vspace{-4mm}
\end{figure}

\section{Discussion}
\subsection{Analysis of generated EEG features}
To assess the quality of the generated EEG features $\text{PTDA}(O_D)$ prior to their integration with the real feature set $X$, as detailed in Section~\ref{sec3.2}, we employed Maximum Mean Discrepancy (MMD) \cite{chen2023multi} and t-SNE visualization \cite{van2008visualizing}.
The MMD was used to quantify the distributional similarity between the generated subset $\text{PTDA}(O_D)$ and the real subset $X$ before integration, yielding scores of 0.0357, 0.0436, and 0.0401 for the AD, FTD, and VCI groups, respectively.
These relatively low values indicate that the generated EEG features closely approximate the distribution of real EEG features. 
This is further corroborated by the consistent improvements observed in downstream tasks.

As illustrated in Fig.~\ref{tSNE}, t-SNE visualizations further support these quantitative results. Across all categories, the generated EEG features maintain high spatial proximity to the real EEG features, indicating that the generative model successfully captures the underlying manifolds of EEG.
In particular, the generated EEG features effectively reflect the distribution characteristics observed in each group, while maintaining appropriate diversity. Overall, these results demonstrate that the PTDA produces high-quality EEG features that preserve the statistical properties of real data and enhance sample diversity.

\subsection{Quantitative analysis of GSA}
A sensitivity analysis was conducted to determine the optimal architecture by varying the number of GSA blocks from 1 to 5. As shown in Fig.~\ref{GSA}, increasing the number of blocks from 1 to 3 leads to consistent performance improvements across all evaluation metrics. However, further increasing the depth beyond three blocks results in performance saturation or degradation, with ACC and AUC declining and RMSE increasing, particularly in the multi-class task.
This phenomenon suggests that while deeper networks possess stronger representation capabilities, excessive complexity relative to the limited scale of EEG induces overfitting, thereby compromising generalization ability. Consequently, to balance performance maximization with computational efficiency, we adopt a configuration of $3$ GSA blocks for the final model.

\subsection{Task-guided query attention visualization}
To validate the TGQ module's effectiveness, we visualize the spatial distribution of task-guided representations across EEG channels in Sec.~\ref{sec3.4}. As illustrated in Fig.~\ref{task}, the attention distributions exhibit clear contrasts between the diagnosis and prediction tasks.
The diagnosis branch reveals a localized activation pattern predominantly clustered in the frontal and temporal regions (e.g., Fp1, Fp2, F3, F4, and T3). These areas are closely associated with the characteristic atrophy patterns of AD and FTD, which are important for subtype differentiation.
In contrast, the prediction branch for MMSE prediction exhibits a broader and more diffuse activation pattern, with greater involvement of posterior and occipital regions (e.g., O1 and O2). This indicates that dementia diagnosis depends more on localized pathological features, whereas MMSE prediction requires integrating information from widespread functional networks.
Moreover, it supports the model’s ability to capture meaningful biomarkers for both dementia diagnosis and MMSE
prediction.

\section{Conclusion}
This study proposes the TGSN for dementia diagnosis and MMSE prediction. The MBFF module performs multi-band feature extraction on EEG and fuses them to fully exploit the discriminative information across different bands. The PTDA module expands sample size and diversity based on a diffusion model, effectively mitigating the small-sample problem. The GSA module captures long-range spatial dependencies and dynamic temporal features from EEG representations. The TGQ module mitigates inter-task interference via task-specific feature extraction and achieves dynamic balance and collaborative optimization across both tasks. 
Extensive experiments on the XY02 dataset demonstrate that TGSN significantly outperforms SOTA methods in both dementia diagnosis and MMSE prediction tasks, with external validation on the DS004504 dataset confirming its robust generalization capabilities. 
Additionally, interpretability analyses have identified critical frequency bands associated with distinct dementia subtypes. Overall, the proposed network effectively improves dementia diagnosis and MMSE prediction accuracy under limited EEG. This demonstrates the potential application of multi-task joint optimization in dementia diagnosis.

\bibliographystyle{IEEEtran}
\bibliography{refer}

\end{document}